\documentclass[11pt]{article}

\usepackage[final]{acl}

\usepackage{times}
\usepackage{latexsym}
\usepackage{amsmath}
\usepackage[T1]{fontenc}

\usepackage[utf8]{inputenc}
\usepackage{booktabs}
\usepackage{multirow}
\usepackage{microtype}

\usepackage[linesnumbered,ruled,vlined]{algorithm2e}

\usepackage{inconsolata}

\usepackage{graphicx}

\usepackage{tcolorbox}
\usepackage{xcolor}
\usepackage{soul}
\usepackage{colortbl}
\usepackage{graphicx}
\usepackage{amssymb}
\usepackage{pifont}
\usepackage{caption}
\usepackage{subcaption}
\usepackage{graphicx}
\usepackage{epstopdf}

\definecolor{lightred}{RGB}{255, 245, 245}
\definecolor{lightblue}{RGB}{245, 250, 255}
\definecolor{darkblue}{RGB}{0, 0, 139}
\definecolor{taggray}{RGB}{100, 100, 100}
\definecolor{softred}{RGB}{190, 30, 30}
\definecolor{arrowcolor}{RGB}{80, 80, 80}
\definecolor{processblue}{RGB}{0, 50, 150}
\definecolor{customblue}{RGB}{65, 105, 225}

\newcommand{\cmark}{\ding{51}}

\title{

MedLayBench-V: A Large-Scale Benchmark for Expert-Lay Semantic Alignment in Medical Vision Language Models

}

\usepackage{hyperref}

\hypersetup{
    colorlinks=true,
    linkcolor=darkblue,
    filecolor=magenta,      
    urlcolor=processblue,
    citecolor=darkblue,
}

\author{
  \textbf{Han Jang\textsuperscript{$\clubsuit$,$\triangle$,$\spadesuit$,$\dagger$}}, 
  \textbf{Junhyeok Lee\textsuperscript{$\clubsuit$,$\heartsuit$,$\spadesuit$,$\dagger$}}, 
  \textbf{Heeseong Eum\textsuperscript{$\clubsuit$,$\heartsuit$,$\spadesuit$}}, 
  \textbf{Kyu Sung Choi\textsuperscript{$\clubsuit$,$\heartsuit$,$\triangle$,$\diamondsuit$,$\spadesuit$,*}}
\\
  \small \textsuperscript{$\clubsuit$}Seoul National University \\
  \small \textsuperscript{$\heartsuit$}Seoul National University College of Medicine \\
  \small \textsuperscript{$\triangle$}Department of Radiology, Seoul National University Hospital \\
  \small \textsuperscript{$\diamondsuit$}Healthcare AI Research Institute, Seoul National University Hospital \\
  \small \textsuperscript{$\spadesuit$}The Advanced Imaging and Computational Neuroimaging~(AICON) Laboratory \\
  \small \texttt{\{hanjang, jhlee0619, seong6466\}@snu.ac.kr, ent1127@snu.ac.kr}
\\
  [0.005cm]
\href{https://janghana.github.io/MedLayBench-V/}{%
    \footnotesize \textbf{\textcolor{customblue}{\raisebox{-0.1em}{\includegraphics[height=1.0em]{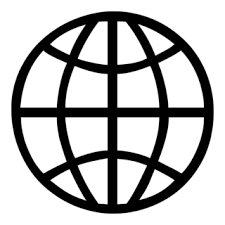}} Project Page}}%
  }
  \hspace{1.0cm}
\href{https://github.com/janghana/MedLayBench-V}{%
    \footnotesize \textbf{\textcolor{customblue}{\raisebox{-0.1em}{\includegraphics[height=1.0em]{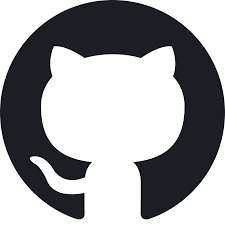}} Code}}%
  }
  \hspace{1.0cm} 
  \href{https://huggingface.co/datasets/hanjang/MedLayBench-V}{%
    \footnotesize \textbf{\textcolor{customblue}{\raisebox{-0.1em}{\includegraphics[height=1.0em]{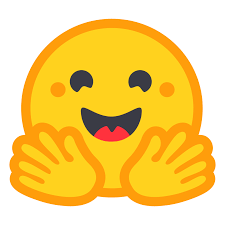}} Dataset}}%
  }
}

\begin{document}
\maketitle

\begingroup
  \renewcommand{\thefootnote}{}
  \footnotetext{\textsuperscript{$\dagger$}These authors contributed equally to this work.}
  \footnotetext{\textsuperscript{*}Corresponding author: \texttt{ent1127@snu.ac.kr}}
\endgroup

\begin{abstract}

Medical Vision-Language Models~(Med-VLMs) have achieved expert-level proficiency in interpreting diagnostic imaging. 
However, current models are predominantly trained on professional literature, limiting their ability to communicate findings in the lay register required for patient-centered care.
While text-centric research has actively developed resources for simplifying medical jargon, there is a critical absence of large-scale multimodal benchmarks designed to facilitate lay-accessible medical image understanding.
To bridge this resource gap, we introduce \textbf{MedLayBench-V}, the first large-scale multimodal benchmark dedicated to expert-lay semantic alignment.
Unlike naive simplification approaches that risk hallucination, our dataset is constructed via a Structured Concept-Grounded Refinement~(SCGR) pipeline.
This method enforces strict semantic equivalence by integrating Unified Medical Language System~(UMLS) Concept Unique Identifiers~(CUIs) with micro-level entity constraints.
MedLayBench-V provides a verified foundation for training and evaluating next-generation Med-VLMs capable of bridging the communication divide between clinical experts and patients.

\end{abstract}

\section{Introduction}

Enhancing the linguistic accessibility of clinical documentation has emerged as a paramount objective in biomedical Natural Language Processing~(NLP).
Driven by the imperative to facilitate patient-centered care, recent research has coalesced around tasks such as Biomedical Lay Summarization~(BioLaySumm) and Neural Text Simplification~(NTS)~\cite{shardlow2019neural, yao2024readme}.
Collectively framed as Medical Lay Language Generation~(MLLG), these efforts aim to translate highly specialized medical jargon into the accessible lay register.
This paradigm shift is epitomized by initiatives like the BioLaySumm shared tasks~\cite{xiao2025overview, goldsack2024overview} and recent benchmarks like MedAgentBoard~\cite{zhu2025medagentboard}, where MLLG is established as a core competency for medical artificial intelligence~(AI).
Recent studies attribute success in this domain to the advanced semantic reasoning of Large Language Models~(LLMs), which allows them to modify lexical complexity while maintaining semantic invariance, thereby ensuring that core medical facts are preserved despite the stylistic shift~\cite{liao2025magical}.

\begin{figure}[t]
    \centering
    \includegraphics[width=\columnwidth]{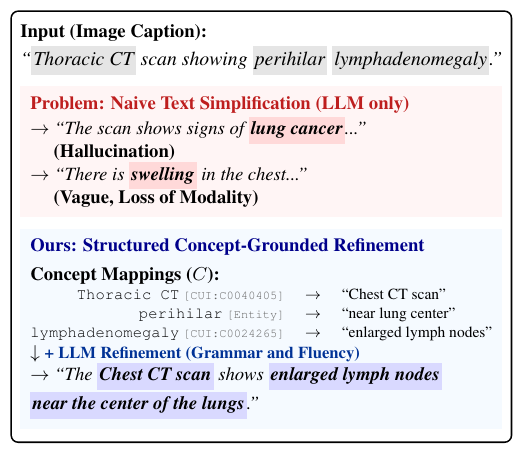}
    \caption{
    \textbf{Motivation.} 
     Our method prevents hallucinations by enforcing Structured Constraints: It explicitly maps extracted Concepts and Entities (e.g., \textit{lymphadenomegaly}) to lay terms, ensuring diagnostic accuracy while preserving specific details.
    }
    \label{fig:motivation}
\end{figure}

While the text-to-text simplification landscape has advanced significantly, the integration of this lay perspective into multimodal systems remains an open challenge.
Medical Vision-Language Models~(Med-VLMs), such as those trained on ROCOv2~\cite{ruckert2024rocov2} or PMC-OA, have achieved expert-level proficiency in interpreting diagnostic imaging~\cite{lozano2025biomedica}.
However, a critical limitation persists in their current training paradigm.
Unlike text-centric LLMs that are becoming increasingly adaptable to the lay register, current Med-VLMs are predominantly optimized for the rigid clinical jargon found in professional literature.
As illustrated in Figure~\ref{fig:motivation}, this domain-specific optimization creates a significant barrier to usability; while models successfully encode visual features into technical tokens like `Pneumothorax', their ability to ground the same visual evidence in natural language equivalents like `Collapsed lung' remains unsupported due to the lack of parallel lay data.
This suggests that without a dedicated benchmark to facilitate expert-to-lay alignment, Med-VLMs will remain confined to a specialized lexicon, severely limiting their applicability in patient-centered care.

Overcoming this resource scarcity, however, presents significant methodological challenges.
Existing multimodal benchmarks are exclusively populated with expert-level reports and offer no ground truth for lay-accessible descriptions.
Furthermore, relying on standard lexical metrics like BLEU~\cite{papineni2002bleu} is insufficient for validation as they inherently penalize the vocabulary shifts required for simplification~\cite{zhao2024x}.
Moreover, constructing a benchmark via naive LLM generation carries the risk of hallucination or the omission of vital quantitative details, which compromises the factual integrity required for medical AI~\cite{liao2025magical}.

To bridge this divide, we introduce \textbf{MedLayBench-V}, the first multimodal benchmark designed to facilitate patient-centric medical image understanding.
Drawing inspiration from recent text-centric approaches that leverage structured medical knowledge to enhance summary relevance~\cite{ming2025towards}, we extend this philosophy to the multimodal domain via a novel \textbf{Structured Concept-Grounded Refinement~(SCGR)} pipeline.
Our approach synergizes macro-level conceptual mapping from the Unified Medical Language System~(UMLS) with micro-level entity constraints extracted via Named Entity Recognition~(NER)~\cite{bodenreider2004unified}.
This hybrid strategy ensures that the generated lay captions maintain strict semantic equivalence with the original expert reports while effectively transitioning to the lay register.
Using this verified dataset, we establish the first comprehensive baselines for expert-lay alignment, providing a standardized foundation for future research in accessible medical AI.

Our contributions are summarized as follows:
\begin{itemize}
    \item To the best of our knowledge, we introduce \textbf{MedLayBench-V}, the first foundational benchmark encompassing diverse medical imaging modalities specifically curated to bridge the linguistic divide between clinical experts and laypersons.
    \item We propose the SCGR pipeline, a verifiable framework that extends knowledge-guided text simplification principles to vision-language tasks, ensuring high clinical correctness and hallucination control.
    \item We establish a comprehensive evaluation protocol for Expert-Lay semantic alignment and provide standardized baselines, offering a robust foundation for future research in patient-centered medical AI.
\end{itemize}

\section{Related Works}

\subsection{Patient-Centered Clinical Reporting}
The complexity of medical documentation creates significant barriers to patient understanding, driving the need for automated systems that can translate clinical narratives into accessible language.
To address this, the field has evolved from early Neural Text Simplification~(NTS) efforts into the broader paradigm of Medical Lay Language Generation~(MLLG)~\cite{shardlow2019neural, yao2024readme}.
This transition is marked by large-scale community initiatives such as the BioLaySumm shared tasks and the MedAgentBoard benchmark, which provide standardized tasks to bridge the communication gap between experts and laypersons~\cite{xiao2025overview, zhu2025medagentboard}.

Within this text-centric landscape, LLMs have achieved remarkable proficiency, effectively balancing lexical simplification with semantic invariance as demonstrated by frameworks~\cite{liao2025magical}.
However, this progress has yet to permeate the multimodal domain.
Unlike the thriving domain for text-only models, there is a critical absence of benchmarks designed to evaluate Med-VLMs leaving it unclear whether current SOTA models can successfully ground visual findings in lay-accessible language without compromising factual accuracy.

\subsection{Medical Vision-Language Models and Dataset Scarcity}
In the multimodal domain, Med-VLMs have achieved expert-level proficiency in interpreting diagnostic imaging~\cite{zhang2023biomedclip, li2023llava, sellergren2025medgemma}.
These capabilities are predominantly driven by large-scale datasets such as ROCOv2~\cite{ruckert2024rocov2} and BIOMEDICA~\cite{lozano2025biomedica}.
However, these datasets are exclusively curated from professional biomedical literature, thereby optimizing models strictly for the rigid clinical jargon.

A critical limitation in existing multimodal datasets is the scarcity of parallel multimodal data that pairs medical images with patient-friendly descriptions.
While models can successfully align visual features with technical concepts (e.g., ``Pneumothorax''), the lack of ground truth for natural language equivalents (e.g., ``Collapsed lung'') prevents them from learning the lay register.
Unlike the text domain where lay benchmarks exist, the vision-language field suffers from this fundamental resource gap, which hinders the development of expert-lay alignment capabilities in VLMs.

\begin{figure*}[t]
    \centering
    \includegraphics[width=\textwidth]{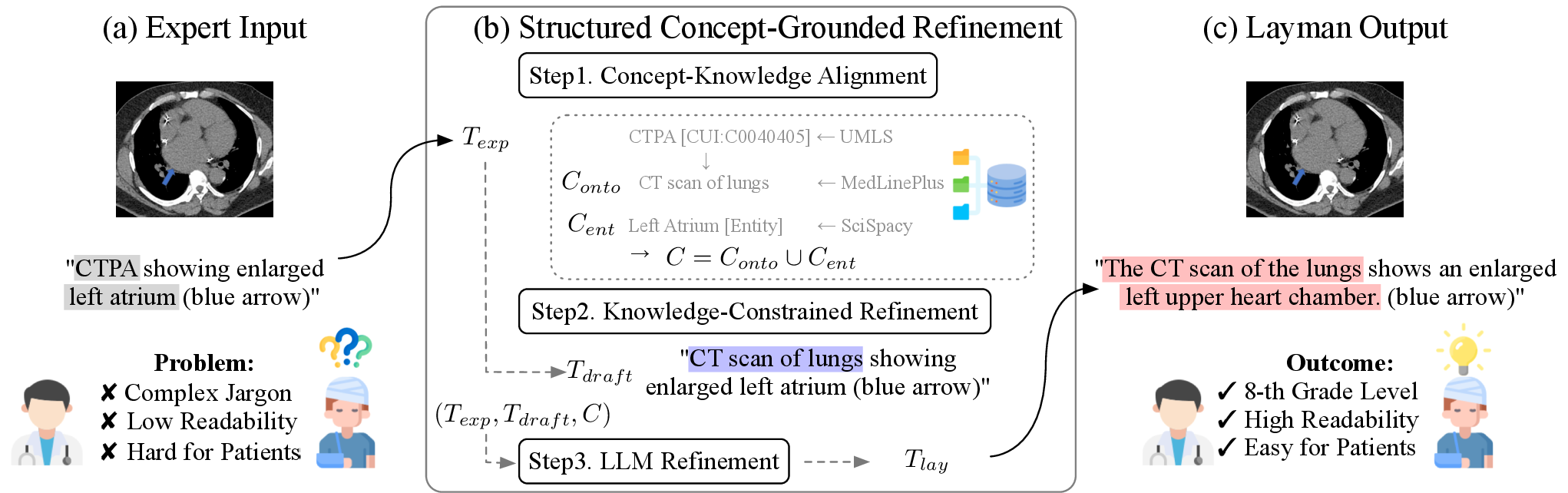}
    \caption{\textbf{Overview of the SCGR Framework.} 
    (a) Expert Input extracts technical concepts from the initial jargon-heavy reports. 
    (b) Structured Concept-Grounded Refinement maps terms to lay definitions and employs Llama-3.1-8B~\cite{dubey2024llama} to synthesize the final caption, optimizing for syntax and fluency while strictly adhering to factual constraints (Detailed prompt in Appendix~\ref{sec:appendix_prompts}).
    (c) Layman Output provides a clinically accurate and accessible description.
    }
    \label{fig:pipeline}
\end{figure*}

\subsection{Limitations of Current Benchmarks}
To bridge the expert-lay divide, prior research has predominantly focused on text-to-text simplification strategies.
Early approaches relied on rule-based methods or phrase tables to substitute medical jargon with simpler synonyms~\cite{shardlow2019neural}.
With the advent of LLMs, recent studies have shifted towards generative rewriting, employing models such as GPT-4o to translate clinical notes into patient-friendly language~\cite{yao2024readme}.
However, LLMs frequently generate plausible yet factually incorrect descriptions or omit vital quantitative details to satisfy readability constraints, thereby compromising patient safety in clinical settings~\cite{moor2023foundation,zhu2025can}.
For instance, a recent prospective trial demonstrated that while LLM-based simplification significantly reduces cognitive workload, it introduced factual errors and omissions in approximately 6--7\% of reports, necessitating rigorous verification mechanisms~\cite{prucker2025prospective}.

Recent initiatives, such as the BioLaySumm 2025 Shared Task~\cite{goldsack2022making,xiao2025overview} and Layman's RRG~\cite{zhao2024x}, have begun to incorporate visual modalities to address these grounding issues.
Despite these advances, current multimodal benchmarks remain limited in scope, predominantly focusing on specific modalities like Chest X-rays~(CXR) with restricted dataset sizes.
Furthermore, these datasets typically rely on end-to-end LLM generation for creating lay captions, which can perpetuate the very hallucinations they aim to resolve without rigorous concept-level verification.
To facilitate the training of robust, general-purpose Med-VLMs, there is a critical need for a large-scale, diverse benchmark that extends beyond single modalities.

\subsection{Evaluation Metrics for Medical Text Generation}
Evaluating the quality of MLLG systems remains a persistent challenge due to the inadequacy of existing metrics.
Traditional n-gram based metrics such as BLEU~\cite{papineni2002bleu}, ROUGE~\cite{lin2004rouge}, and METEOR~\cite{banerjee2005meteor} measure surface-level overlap.
However, they inherently penalize the vocabulary shifts required for simplification, making them unsuitable for expert-to-lay translation tasks~\cite{zhao2024x, zhang2019bertscore}.
Conversely, medically-oriented metrics like Green~\cite{ostmeier2024green} and RaTEScore~\cite{zhao2024ratescore} focus on clinical factuality and entity extraction.

While effective for expert reports, they do not assess whether the generated text is understandable to a lay audience.
Finally, standard readability metrics rely on heuristic formulas (e.g., sentence length) rather than actual comprehensibility, often failing to capture the semantic nuances required for patient education~\cite{yao2024readme}.
Therefore, effective MLLG evaluation requires a comprehensive framework that simultaneously assesses visual grounding, factual correctness, and lay accessibility.
However, performing such multi-dimensional evaluation is unfeasible with current VLM datasets due to the critical absence of lay-aligned references.
To bridge this gap, we introduce MedLayBench-V, a unified benchmark designed to facilitate this holistic evaluation.

\section{Methodology}
\label{sec:method}

We introduce \textbf{MedLayBench-V}, a large-scale multimodal benchmark designed to bridge the gap between expert clinical jargon and patient-accessible language.
To ensure the high semantic fidelity of this benchmark, we propose the Structured Concept-Grounded Refinement (SCGR) pipeline.
Crucially, our framework explicitly decouples semantic extraction from stylistic refinement.
This separation ensures strict \textit{Semantic Equivalence} between the expert and lay registers, mitigating the hallucinations common in end-to-end generation.
The pipeline consists of three distinct stages, \textbf{corresponding to Steps 1--3 in Figure~\ref{fig:pipeline}(b)}: (i) Concept-Knowledge Alignment, (ii) Knowledge-Constrained Refinement, and (iii) LLM Refinement.

\subsection{Data Source and Task Definition}
We utilize the ROCOv2 dataset~\cite{ruckert2024rocov2}\footnote{\url{https://huggingface.co/datasets/eltorio/ROCOv2-radiology}} as our seed corpus.
Derived from the PubMed Central Open Access (PMC-OA) subset~\cite{lin2023pmc}\footnote{\url{https://pmc.ncbi.nlm.nih.gov/tools/openftlist/}}, ROCOv2 is uniquely advantageous for our task as it provides not only diagnostic captions ($T_{exp}$) but also pre-computed UMLS Concept Unique Identifiers (CUIs) extracted via the MedCAT toolkit~\cite{kraljevic2021multi}\footnote{\url{https://github.com/CogStack/MedCAT2}}.
These existing annotations serve as a critical foundation for our semantic extraction phase.

Despite the richness of this clinical metadata, the expert descriptions in ROCOv2 remain inherently unintelligible to non-specialists.
Our objective is to augment these pairs with layman-accessible descriptions ($T_{lay}$), creating the first dual-register medical benchmark optimized for patient-centric VLM training and testing.

\subsection{Concept-Knowledge Alignment}
To guarantee that the simplified captions retain the diagnostic precision of $T_{exp}$, we first extract a set of semantic constraints $C$.
This process integrates high-level ontology mapping with fine-grained entity recognition.

\begin{table*}[t]
\centering
\caption{\textbf{Linguistic Complexity and Readability Analysis.} 
Our refinement consistently reduces reading difficulty, improves accessibility, and standardizes vocabulary across the entire dataset.}
\label{tab:linguistic_full}
\resizebox{\textwidth}{!}{
\begin{tabular}{l|cc|cc|cc|cc}
\toprule
\multirow{3}{*}{\textbf{Linguistic Metric}} & \multicolumn{2}{c|}{\textbf{Train Set}} & \multicolumn{2}{c|}{\textbf{Validation Set}} & \multicolumn{2}{c|}{\textbf{Test Set}} & \multicolumn{2}{c}{\textbf{Overall (Total)}} \\
 & \multicolumn{2}{c|}{($N=59,962$)} & \multicolumn{2}{c|}{($N=9,904$)} & \multicolumn{2}{c|}{($N=9,927$)} & \multicolumn{2}{c}{($N=79,793$)} \\
\cmidrule(lr){2-3} \cmidrule(lr){4-5} \cmidrule(lr){6-7} \cmidrule(lr){8-9}
 & \textbf{Expert} & \textbf{Layman} & \textbf{Expert} & \textbf{Layman} & \textbf{Expert} & \textbf{Layman} & \textbf{Expert} & \textbf{Layman} \\
\midrule
\multicolumn{9}{l}{\textit{\textbf{Readability Metrics}}} \\
\quad FKGL~\cite{kincaid1975derivation} $\downarrow$ & 13.05 & \textbf{10.29} & 13.29 & \textbf{10.50} & 13.21 & \textbf{10.53} & 13.10 & \textbf{10.35} \\
\quad CLI~\cite{coleman1975computer} $\downarrow$ & 15.73 & \textbf{9.82} & 16.12 & \textbf{10.06} & 16.02 & \textbf{10.04} & 15.82 & \textbf{9.88} \\
\quad DCRS~\cite{dale1948formula} $\downarrow$ & 14.02 & \textbf{11.73} & 14.09 & \textbf{11.80} & 14.02 & \textbf{11.77} & 14.03 & \textbf{11.74} \\
\quad SMOG~\cite{mc1969smog} $\downarrow$ & 13.71 & \textbf{12.21} & 13.85 & \textbf{12.35} & 13.88 & \textbf{12.41} & 13.75 & \textbf{12.25} \\
\quad FRE~\cite{flesch1948new} $\uparrow$ & 26.44 & \textbf{56.15} & 24.85 & \textbf{55.00} & 25.64 & \textbf{55.09} & 26.14 & \textbf{55.88} \\
\midrule
\multicolumn{9}{l}{\textit{\textbf{Lexical Statistics}}} \\
\quad Average Sentence Length & 23.73 & 27.81 & 25.45 & 28.86 & 25.61 & 29.34 & 24.17 & 28.13 \\
\quad Vocab Size $\downarrow$ & 36,875 & \textbf{20,589} & 14,877 & \textbf{9,191} & 14,865 & \textbf{9,238} & 44,673 & \textbf{24,085} \\
\bottomrule
\end{tabular}
}
\end{table*}

\paragraph{Ontology-Based CUI Mapping.}
We utilize the UMLS Metathesaurus API~\cite{bodenreider2004unified}\footnote{\url{https://www.nlm.nih.gov/research/umls/}} to ground clinical terms to CUIs.
In contrast to heuristic string matching, direct API querying guarantees precise alignment with standard medical ontologies.
This step captures core medical concepts (e.g., \texttt{C0040405} $\rightarrow$ ``CTPA'').
We denote the set of identified CUIs as $C_{onto}$, ensuring that the pathology is rigorously anchored to standardized terminology.

\paragraph{Fine-Grained Entity Extraction.}
We supplement CUIs with a biomedical Named Entity Recognition~(NER) model, SciSpacy~\cite{neumann2019scispacy}\footnote{\url{https://allenai.github.io/scispacy/}}.
This module explicitly extracts quantitative attributes (e.g., lesion sizes) and spatial descriptors ($C_{ent}$) often missed by high-level mapping.
We integrate these two sources to establish the final semantic constraint set $C$. Formally, this is defined as:

\begin{equation}
    C = C_{onto} \cup C_{ent}
\end{equation}

\noindent where $C_{onto}$ represents the high-level ontological constraints anchored to UMLS, and $C_{ent}$ denotes the fine-grained entity constraints extracted via NER.

\subsection{Knowledge-Constrained Refinement}
Leveraging the semantic constraint set $C$, we synthesize the lay caption $T_{lay}$.
This phase shifts the linguistic register while strictly adhering to the extracted medical facts.

\paragraph{Lexical Alignment and Draft Synthesis.}
For each concept in $C_{onto}$, we retrieve patient-friendly definitions by querying the MedlinePlus vocabulary within the UMLS Metathesaurus.
Curated by the National Library of Medicine~(NLM), MedlinePlus serves as the authoritative bridge between rigorous clinical ontologies and public health literacy~\cite{miller2000medlineplus}\footnote{\url{https://medlineplus.gov/}}.
By aligning UMLS CUIs directly with MedlinePlus definitions, we ensure that the terminology is not merely simplified but standardized to a trusted lay register.
We then construct an intermediate noisy lay draft ($T_{draft}$) via deterministic dictionary-based substitution.
While grammatically noisy, $T_{draft}$ serves as a reliable lexical basis for the subsequent refinement.

\begin{figure*}[t]
    \centering
    \includegraphics[width=\textwidth]{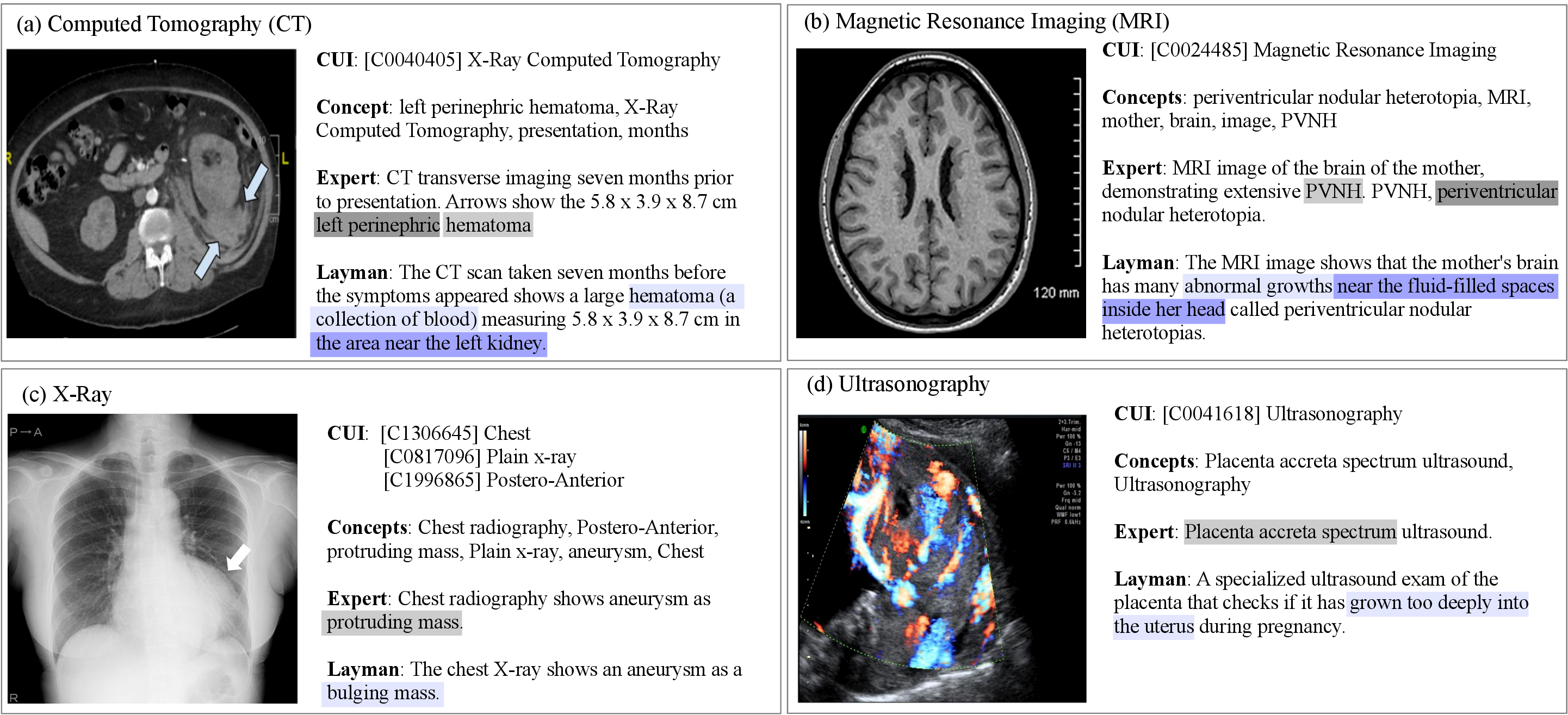}
    \caption{
    \textbf{Qualitative Comparison of Jargon Refinement across Modalities.}
    The figure illustrates example cases from CT, MRI, X-Ray, and Ultrasound.
    Highlights indicate the transformation from medical jargon (Original expert-level caption) to patient-friendly language (Layman-level caption).
    Our method successfully simplifies anatomical terms, structural definitions, and visual descriptions while preserving core medical information. Additional examples are provided in Appendix~\ref{sec:appendix_qualitative}.
    }
    \label{fig:qualitative_results}
\end{figure*}

\paragraph{Constraint-Guided Linguistic Refinement.}
To generate the final accessible caption, we employ Llama-3.1-8B-Instruct~\cite{dubey2024llama} within a constrained generation framework.
We chose Llama-3.1-8B-Instruct for this stage due to its open-weight reproducibility, computational practicality for processing approximately 80K samples, and strong instruction-following capability for constrained text refinement.
Since the structured constraints are responsible for preserving semantic fidelity, the LLM's role is limited to grammar and fluency optimization, which does not require a larger model or domain-specific medical knowledge.
Our structured prompt incorporates: (1) the source text $T_{exp}$ ensuring factual grounding, (2) a strict constraint set $C$ for hallucination mitigation, and (3) the initial draft $T_{draft}$ to steer vocabulary selection.
The objective is to downscale linguistic complexity from a college-level register to a high school level, ensuring the output remains semantically faithful to the clinical findings through explicit constraints.
Figure~\ref{fig:qualitative_results} demonstrates qualitative examples of our refinement across different modalities.

\begin{table}[t]
\centering
\caption{\textbf{Dataset Statistics and Quality Consistency.} 
We report consistency across Train ($N$=59,962), Validation ($N$=9,904), and Test ($N$=9,927). The \textbf{Overall} column represents the weighted average ($N$=79,793). High clinical correctness (RaTEScore, GREEN) and consistent simplification scores (LENS) across all splits confirm the robust quality of our refinement pipeline.}
\label{tab:split_quality_single}
\resizebox{\columnwidth}{!}{
\setlength{\tabcolsep}{4pt}
\renewcommand{\arraystretch}{1.2}
\begin{tabular}{l|ccc|c}
\toprule
\textbf{Metric} & \textbf{Train} & \textbf{Val} & \textbf{Test} & \textbf{Overall} \\
\midrule
\multicolumn{5}{l}{\textit{\textbf{Relevance}}} \\
\quad BLEU-4~\cite{papineni2002bleu} $\uparrow$ & 20.99 & 22.32 & 22.45 & \textbf{21.34} \\
\quad ROUGE-L~\cite{lin2004rouge} $\uparrow$ & 49.33 & 50.13 & 50.40 & \textbf{49.56} \\
\quad METEOR~\cite{banerjee2005meteor} $\uparrow$ & 53.00 & 53.40 & 53.56 & \textbf{53.12} \\
\midrule
\multicolumn{5}{l}{\textit{\textbf{Readability}}} \\
\quad LENS~\cite{maddela2023lens} $\uparrow$ & 63.28 & 62.91 & 62.94 & \textbf{63.19} \\
\midrule
\multicolumn{5}{l}{\textit{\textbf{Radiological Factuality}}} \\
\quad RaTEScore~\cite{zhao2024ratescore} $\uparrow$ & 64.66 & 64.57 & 65.09 & \textbf{64.70} \\
\quad GREEN~\cite{ostmeier2024green} $\uparrow$ & 69.03 & 70.14 & 70.03 & \textbf{69.29} \\
\bottomrule
\end{tabular}
}
\end{table}

\section{Experiments}
\label{sec:experiments}

We demonstrate the value of \textbf{MedLayBench-V} through a comprehensive analysis of its linguistic properties and quality consistency, followed by a zero-shot downstream benchmark to evaluate current VLMs' capability in handling both expert and layman medical concepts.

\subsection{Evaluation Metrics}
\label{subsec:metrics}

To ensure a comprehensive assessment, we employ metrics across four dimensions: textual similarity, linguistic readability, clinical factuality, and downstream utility.

\begin{itemize}
    \item \textbf{Relevance:} We use standard n-gram metrics to measure the structural similarity and lexical overlap between expert and layman captions. Specifically, we report BLEU-4~\cite{papineni2002bleu}, ROUGE-L~\cite{lin2004rouge}, and METEOR~\cite{banerjee2005meteor}.

    \item \textbf{Readability:} To quantify the accessibility of the text, we utilize Flesch-Kincaid Grade Level~(FKGL)~\cite{kincaid1975derivation}, Coleman-Liau Index~(CLI)~\cite{coleman1975computer}, Dale-Chall Readability Score~(DCRS)~\cite{dale1948formula}, Simple Measure of Gobbledygook~(SMOG) Index~\cite{mc1969smog}, and Flesch Reading Ease~(FRE)~\cite{flesch1948new}. Additionally, we incorporate LENS~\cite{maddela2023lens}, a learnable metric specifically optimized for text simplification.

    \item \textbf{Radiological Factuality:} Evaluating the clinical integrity of simplified text is critical. We employ Radiological Report Text Evaluation~(RaTEScore)~\cite{zhao2024ratescore} and Generative Radiology Report Evaluation and Error Notation~(GREEN)~\cite{ostmeier2024green}. These model-based metrics are designed to detect hallucinations and ensure clinical correctness in radiology reports.

    \item \textbf{Downstream Performance:} To assess whether the simplified text preserves essential semantic information for automated analysis, we evaluate zero-shot text-to-image retrieval performance. We report Recall@K (R@1, R@5, R@10) to measure retrieval accuracy using the generated captions.
\end{itemize}

\begin{table*}[t!]
\centering
\caption{Overall top-K retrieval performance on MedLayBench-V across four modalities (X-Ray, CT, MRI, Ultrasound). \textbf{Bold} indicates best performance, \underline{underline} indicates second best performance. Values are presented as Expert / Layman. All values are in percentage (\%).}
\label{tab:retrieval_performance}
\resizebox{\textwidth}{!}{
\setlength{\tabcolsep}{1pt}
\renewcommand{\arraystretch}{1.0}
\begin{tabular}{l ccc ccc}
\toprule
\multirow{2}{*}{Model} & \multicolumn{3}{c}{Image $\rightarrow$ Text} & \multicolumn{3}{c}{Text $\rightarrow$ Image} \\ 
\cmidrule(lr){2-4} \cmidrule(lr){5-7}
 & Recall@1 & Recall@5 & Recall@10 & Recall@1 & Recall@5 & Recall@10 \\ 
\midrule

OpenAI-Base & 1.23 / 1.08 & 3.96 / 3.74 & 6.56 / 6.32 & 1.57 / 1.54 & 4.41 / 4.51 & 7.03 / 7.12 \\
CoCa-Large & 2.10 / 2.15 & 5.70 / 5.71 & 8.24 / 8.07 & 3.56 / 3.64 & 8.78 / 8.84 & 11.97 / 12.11 \\
LAION-2B & 2.28 / 2.33 & 6.67 / 6.58 & 9.88 / 9.78 & 4.31 / 4.29 & 9.94 / 9.88 & 13.76 / 13.74 \\
OpenCLIP-Huge & 3.33 / 3.44 & 8.71 / 8.43 & 12.58 / 12.28 & 5.17 / 5.15 & 11.88 / 12.10 & 16.59 / 16.70 \\ 
\midrule

PubMedCLIP & 4.61 / 4.26 & 13.46 / 13.12 & 20.93 / 20.66 & 4.85 / 4.71 & 14.49 / 14.43 & 21.94 / 21.73 \\
BMC-CLIP & 22.69 / 22.42 & 40.83 / 40.36 & 50.33 / 49.65 & 23.04 / 23.21 & 42.09 / 42.03 & 52.09 / 51.71 \\
PMC-CLIP & \underline{28.98} / \underline{28.38} & \underline{53.12} / \underline{52.47} & \underline{64.14} / \underline{63.60} & \underline{30.90} / \underline{30.24} & \underline{55.66} / \underline{55.16} & \underline{66.11} / \underline{65.55} \\
BiomedCLIP & \textbf{31.06} / \textbf{30.70} & \textbf{58.52} / \textbf{58.11} & \textbf{70.31} / \textbf{69.41} & \textbf{32.50} / \textbf{32.07} & \textbf{59.94} / \textbf{59.09} & \textbf{71.07} / \textbf{70.44} \\ 

\bottomrule
\end{tabular}
}
\end{table*}

\subsection{Dataset Statistics and Quality Analysis}
\label{subsec:statistics}

We analyze the linguistic characteristics and semantic consistency of MedLayBench-V, which comprises 79,789 image-text pairs across 7 modalities, maintaining the original ROCOv2 configuration~\cite{ruckert2024rocov2}.

\paragraph{Linguistic Complexity and Accessibility.}
As presented in Table~\ref{tab:linguistic_full}, our refinement pipeline successfully standardizes the linguistic complexity of medical captions.

\begin{itemize}
    \item \textbf{Vocabulary Reduction:} The unique vocabulary size is reduced by 46.1\% in the layman version compared to the expert version. This indicates a significant removal of long-tail medical jargon and noisy tokens, streamlining the dataset for generalizable learning.
    
    \item \textbf{Improved Readability:} We observe a consistent drop in grade-level metrics across the entire dataset. Notably, the FKGL drops from 13.10 to 10.35, and the Coleman-Liau Index decreases from a graduate level of 15.82 to 9.88, aligning with the recommended reading level for patient education materials~\cite{rooney2021readability}.
    
    \item \textbf{Enhanced Accessibility:} The FRE score more than doubles from 26.14 to 55.88. This shift in text difficulty from very confusing to fairly difficult ensures the content is accessible to a general audience with a standard high school education.
\end{itemize}

Detailed modality distributions and concept frequency analyses are provided in Appendix~\ref{sec:appendix_dataset}.

\paragraph{Quality Consistency across Splits.}
Table~\ref{tab:split_quality_single} reports the semantic quality and consistency of our dataset.
The relevance metrics, including BLEU-4, ROUGE-L, and METEOR, show minimal variance across training, validation, and test sets, with an overall METEOR of 53.12, confirming that our pipeline produces stylistically consistent outputs regardless of data split.
The LENS score, a learnable metric for text simplification, remains stable at 63.19 across all splits, indicating robust rewriting quality throughout the dataset.
Most importantly, the clinical correctness scores, RaTEScore and GREEN, demonstrate that our simplification preserves the factual integrity of the original reports, with the test set achieving 65.09 and 70.03 respectively, confirming high clinical safety despite reduced linguistic complexity.

    
    

\subsection{Human Evaluation}
\label{subsec:human_eval}

To validate the SCGR-generated captions beyond automatic metrics, we conducted a human evaluation following~\citet{jeblick2024chatgpt}.
Two board-certified radiologists and one lay reader rated 100 randomly sampled caption pairs on a 5-point Likert scale across four criteria: Factual Correctness, Completeness, Simplicity, and Fluency.

\begin{table}[t]
\centering
\caption{\textbf{Human Evaluation Results.}
Two radiologists (E1, E2) and one lay reader (L) rated 100 SCGR-generated caption pairs on a 5-point Likert scale.}
\label{tab:human_eval}
\resizebox{\columnwidth}{!}{
\setlength{\tabcolsep}{4pt}
\renewcommand{\arraystretch}{1.2}
\begin{tabular}{l|ccc|c}
\toprule
\textbf{Criterion} & \textbf{E1} & \textbf{E2} & \textbf{L} & \textbf{Avg.} \\
\midrule
Factual Correctness & 4.67{\scriptsize $\pm$0.68} & 4.96{\scriptsize $\pm$0.40} & 4.95{\scriptsize $\pm$0.17} & 4.86 \\
Completeness & 4.66{\scriptsize $\pm$0.68} & 4.96{\scriptsize $\pm$0.40} & 4.95{\scriptsize $\pm$0.22} & 4.86 \\
Simplicity & 4.69{\scriptsize $\pm$0.63} & 4.73{\scriptsize $\pm$0.53} & 4.54{\scriptsize $\pm$0.62} & 4.65 \\
Fluency & 4.85{\scriptsize $\pm$0.38} & 4.98{\scriptsize $\pm$0.14} & 4.96{\scriptsize $\pm$0.20} & 4.93 \\
\bottomrule
\end{tabular}
}
\end{table}

As shown in Table~\ref{tab:human_eval}, all criteria averaged above 4.5, with Factual Correctness and Completeness reaching 4.86, confirming that SCGR preserves clinical integrity.
Simplicity scored comparatively lower at 4.65, suggesting room for optimization in certain specialized descriptions.

\subsection{Downstream Task: Zero-Shot Retrieval}
\label{subsec:retrieval}

To evaluate the utility of MedLayBench-V, we conducted a zero-shot Image-Text Retrieval (ITR) experiment.
This task measures how well models can align visual features with both \textit{Expert} (original) and \textit{Layman} (refined) textual descriptions.
We report the Recall@$K$ metrics for both Image-to-Text and Text-to-Image retrieval in Table~\ref{tab:retrieval_performance}, with visualizations provided in Figure~\ref{fig:retrieval_graph} in Appendix~\ref{sec:ablation_llm_only}.
Bootstrap significance testing ($n$=1{,}000, two-sided) confirms that all absolute performance differences remain below 1.03\%, with detailed results in Appendix~\ref{sec:appendix_bootstrap}.
An embedding-level analysis with t-SNE visualizations further confirms this finding (Appendix~\ref{sec:appendix_semantic}).

\paragraph{Experimental Setup.}
Following the standard zero-shot retrieval protocol~\cite{radford2021learning}, we extract image and text embeddings from each dual-encoder model, apply L2-normalization, and compute pairwise cosine similarity across all image-text pairs in the test set ($N$=9,927).
Recall@$K$ is computed by checking whether the ground-truth match appears within the top-$K$ ranked candidates.
No fine-tuning or prompt engineering is applied; all models are evaluated using their publicly available pre-trained weights.

\paragraph{Baseline Models.}
We benchmarked diverse dual-encoder architectures, categorized into general-domain and medical-domain models.
For the general domain, we employed OpenAI-CLIP~\cite{radford2021learning} and OpenCLIP~\cite{cherti2023reproducible} (trained on LAION-2B~\cite{schuhmann2022laion}), along with CoCa~\cite{yu2022coca}, which integrates contrastive and generative objectives.
For the medical domain, we selected models pre-trained on large-scale biomedical image-text pairs to assess the impact of domain adaptation.
These include PubMedCLIP~\cite{eslami2023pubmedclip}, BMC-CLIP~\cite{lozano2025biomedica}, PMC-CLIP~\cite{lin2023pmc}, and BiomedCLIP~\cite{zhang2023biomedclip}, which utilize domain-specific encoders aligned with biomedical imagery.

\paragraph{Performance of Medical vs. General VLMs.}
We observe a clear performance hierarchy based on domain adaptation.
While general domain models (e.g., OpenAI-CLIP) struggle with medical contexts (Recall@1 $< 5\%$), medical-specific models show improved alignment.
BiomedCLIP achieves state-of-the-art performance, benefiting from its large-scale pre-training on biomedical literature.

\paragraph{Semantic Preservation in Layman Captions.}
Crucially, our results demonstrate that simplifying the language does not compromise semantic fidelity.
As evidenced in Table~\ref{tab:retrieval_performance}, retrieval performance remains robust across all medical models, exhibiting negligible degradation when transitioning from \textit{Expert} to \textit{Layman} queries.
For instance, BiomedCLIP exhibits only a marginal drop in Image-to-Text Recall@1 (31.06\% $\rightarrow$ 30.70\%).
This explicitly verifies that MedLayBench-V successfully retains the core diagnostic semantics required for visual alignment, proving high readability can be achieved without sacrificing medical accuracy.

\paragraph{Ablation: Impact of Structured Grounding.}
To isolate each SCGR component, we conducted a systematic ablation (Table~\ref{tab:ablation_scgr}).
Without structured grounding, LLM Only collapses to 1.96 avg R@1, an 83\% drop from Expert.
CUI extraction alone yields negligible recovery, while full SCGR restores 98.4\% of Expert-level performance, confirming knowledge-constrained refinement as the critical component.
Per-model breakdown is provided in Appendix~\ref{sec:appendix_ablation_full}.

\begin{table}[t]
\centering
\caption{\textbf{SCGR Ablation Study.}
Averaged R@1 (\%) across I2T and T2I. Full per-model results in Appendix~\ref{sec:appendix_ablation_full}.}
\label{tab:ablation_scgr}
\resizebox{\columnwidth}{!}{
\setlength{\tabcolsep}{5pt}
\renewcommand{\arraystretch}{1.2}
\begin{tabular}{l|ccc|c}
\toprule
\textbf{Condition} & \textbf{CUI} & \textbf{MedLP} & \textbf{LLM} & \textbf{Avg. R@1} \\
\midrule
LLM Only & $\times$ & $\times$ & \cmark & 1.96 \\
LLM + CUI  & \cmark & $\times$ & \cmark & 2.08 \\
SCGR (Ours) & \cmark & \cmark & \cmark & \textbf{11.26} \\
\midrule
Expert & -- & -- & -- & 11.44 \\
\bottomrule
\end{tabular}
}
\end{table}

\subsection{Downstream Task: Zero-Shot Captioning}
To further expose the expert-lay register gap, we conducted a zero-shot captioning experiment using both medical and general-domain VLMs (Appendix~\ref{sec:appendix_captioning}).
In particular, LLaVA-Med~\cite{li2023llava} exhibits severe expert bias with a BERTScore gap of +22.93 between expert and layman prompts, while other models show near-zero gaps, confirming that lay-register adaptability varies significantly across model families.

\section{Conclusion}
\vspace{-0.5em}
In this work, we introduced MedLayBench-V, the first multimodal benchmark for quantifying the semantic alignment between clinical jargon and lay language.
By evaluating state-of-the-art VLMs, we formalized the existence of a representation alignment gap, revealing that current medical models are overfitted to the professional register at the expense of patient accessibility.
Our proposed structured concept-grounded refinement pipeline provides a foundational framework for developing next-generation Medical AI that is both clinically accurate and universally understandable.

\section*{Acknowledgements}
This work was supported by the National Research Foundation of Korea (NRF) grant funded by the Korea government (MSIT) (No. RS-2023-00251022) (K.S.C.); the SNUH Research Fund (No. 04-2024-0600; No. 04-2025-2060) (K.S.C.); and the Korea Health Technology R\&D Project through the Korea Health Industry Development Institute (KHIDI) grant funded by the Ministry of Health\&Welfare (No. RS-2024-00439549) (K.S.C.).

\section*{Limitations}

While MedLayBench-V establishes a foundation for patient-centric AI, we acknowledge limitations regarding the reliance on synthetic data, restriction to English, and modality imbalances inherited from the source. Although our pipeline ensures clinical correctness via structured constraints, synthetic captions may lack the subtle nuances of human-authored text, and validation with diverse patient groups is needed to assess real-world utility.

More importantly, we hypothesize that the representation alignment gap between clinical jargon and lay language may have been obscured by the limited complexity of the current retrieval task. We posit that a distinct gap exists but requires more challenging scenarios to be fully exposed. Consequently, our future work will focus on scaling this benchmark to a wider array of complex downstream tasks. By increasing both the scale and difficulty, we aim to rigorously identify this latent alignment gap and develop robust methodologies to effectively bridge the expert-lay divide.

Finally, while frontier models such as GPT, Gemini, and Claude may already possess expert-to-lay conversion capabilities, evaluating such ability requires a standardized resource with ontology-grounded references.
MedLayBench-V serves this role by providing paired dual-register data for reproducible comparison across model families, analogous to how ImageNet remains a shared evaluation standard beyond its original difficulty level.

Despite these limitations, we believe MedLayBench-V represents a meaningful step toward closing the communication gap between clinical AI and patients, contributing to equitable and accessible healthcare. We encourage the community to extend this benchmark to multilingual settings, additional imaging modalities, and more diverse downstream tasks such as visual question answering and medical report generation.




\bibliography{ref}

\appendix
\clearpage
\renewcommand{\thefigure}{A\arabic{figure}}
\renewcommand{\thetable}{A\arabic{table}}
\setcounter{figure}{0}
\setcounter{table}{0}

\section{Implementation Details and Prompts}
\label{sec:appendix_prompts}

In this section, we provide a comprehensive breakdown of the SCGR pipeline's implementation.
The core of our approach lies in the rigorous separation of semantic extraction and stylistic refinement, as detailed in Algorithm~\ref{algo:scgr_pipeline}.
To ensure that the LLM adheres strictly to clinical facts while simplifying the syntax, we engineered a specific prompt template shown in Figure~\ref{fig:prompt_interface}.
By explicitly defining the system role as a "Medical Text Simplifier" and enforcing a JSON output format, we enable reliable automated parsing at scale.
The "Critical Instructions" block serves as a safeguard against common pitfalls such as hallucinations or the use of subjective pronouns (e.g., "your body"), ensuring the output remains objective and professional.

\begin{algorithm}[h]
\small
\SetAlgoLined
\DontPrintSemicolon
\SetKwInOut{Input}{Input}
\SetKwInOut{Output}{Output}
\Input{Set of Expert Captions $\mathcal{T}_{exp} = \{T_{exp}^{(1)}, ..., T_{exp}^{(N)}\}$}
\Output{Set of Layman Captions $\mathcal{T}_{lay}$}
\BlankLine
Initialize $\mathcal{T}_{lay} \leftarrow \emptyset$ \;
\ForEach{$T_{exp} \in \mathcal{T}_{exp}$}{
    \tcp{Step 1: Hybrid Concept Extraction}
    $C_{onto} \leftarrow \text{ExtractCUIs}(T_{exp})$ \tcp*[r]{MedCAT}
    $C_{ent} \leftarrow \text{ExtractEntities}(T_{exp})$ \tcp*[r]{SciSpacy}
    $C \leftarrow C_{onto} \cup C_{ent}$ \;
    \BlankLine
    \tcp{Step 2: Knowledge Retrieval \& Drafting}
    $T_{draft} \leftarrow T_{exp}$ \;
    \ForEach{$c \in C_{onto}$}{
        $def \leftarrow \text{RetrieveLayDef}(c)$ \tcp*[r]{MedlinePlus}
        $T_{draft} \leftarrow \text{Substitute}(T_{draft}, c, def)$ \;
    }
    \BlankLine
    \tcp{Step 3: Constrained Refinement (LLM)}
    $P \leftarrow \text{ConstructPrompt}(T_{exp}, C, T_{draft})$ \;
    $T_{lay} \leftarrow \text{Generate}(P)$ \tcp*[r]{Llama-3}
    \BlankLine
    \tcp{Step 4: Quality Verification}
    \If{$\text{CheckFactuality}(T_{lay}, T_{exp})$}{
        $\mathcal{T}_{lay}.\text{add}(T_{lay})$ \;
    }
}
\Return $\mathcal{T}_{lay}$ \;

\caption{SCGR framework}
\label{algo:scgr_pipeline}
\end{algorithm}

\begin{figure}[h]
    \centering
    \begin{tcolorbox}[
        colback=white, 
        colframe=black, 
        boxrule=0.8pt, 
        arc=3pt, 
        left=4pt, right=4pt, top=4pt, bottom=4pt,
        title=\textbf{\sffamily SCGR Instruction Prompt Template}, 
        colbacktitle=black, 
        coltitle=white, 
        fonttitle=\bfseries\sffamily
    ]
    \small \sffamily 
    \textbf{[System Role]} \\
    You are a precise \textbf{Medical Text Simplifier}. Rewrite the report for a high school student using the provided Concepts.
    \vspace{0.15cm}
    \textbf{[Critical Instructions]}
    \begin{enumerate}
        \setlength\itemsep{0em} \setlength\parskip{0em}
        \item \textbf{Source of Truth:} Trust the \texttt{Original Caption} completely. Ignore hallucinations in the Draft.
        \item \textbf{Objective Tone:} \textbf{No 'you'/'your'.} Use 'the patient' or 'the body'.
        \item \textbf{Strict Format:} Return \textbf{ONLY} the refined sentence. No "Note:" or explanations.
        \item \textbf{No Hallucinations:} Do not invent words. Keep unclear terms in parentheses.
    \end{enumerate}
    \hrulefill
    \vspace{0.1cm}
    
    \textbf{[User Input Template]} \\
    
    \begin{tabular}{@{}l p{0.75\linewidth}@{}}
        \textbf{Original (Fact):} & "\{Expert Caption ($T_{exp}$)\}" \\
        \textbf{Concepts:} & [\{Verified UMLS Concepts ($C$)\}] \\
        \textbf{Draft (Ref):} & "\{Noisy Layman Draft ($T_{draft}$)\}" \\
    \end{tabular}
    \hrulefill 
    \vspace{0.1cm}
    \textbf{[Structured Output]} \\
    \texttt{\{} \\
    \hspace*{1em} \texttt{"layman\_caption": "The CT scan shows an enlarged heart..."} \\
    \texttt{\}}
    
    \end{tcolorbox}
    \caption{\textbf{Prompt Construction for SCGR.} 
    The prompt enforces strict adherence to the \textit{Original Caption} as the source of truth while utilizing the \textit{Draft} only for stylistic reference. The output is constrained to an objective, third-person tone.}
    \label{fig:prompt_interface}
\end{figure}

\section{Detailed Dataset Statistics}
\label{sec:appendix_dataset}

\begin{figure*}[h]
    \centering
    \includegraphics[width=0.95\textwidth]{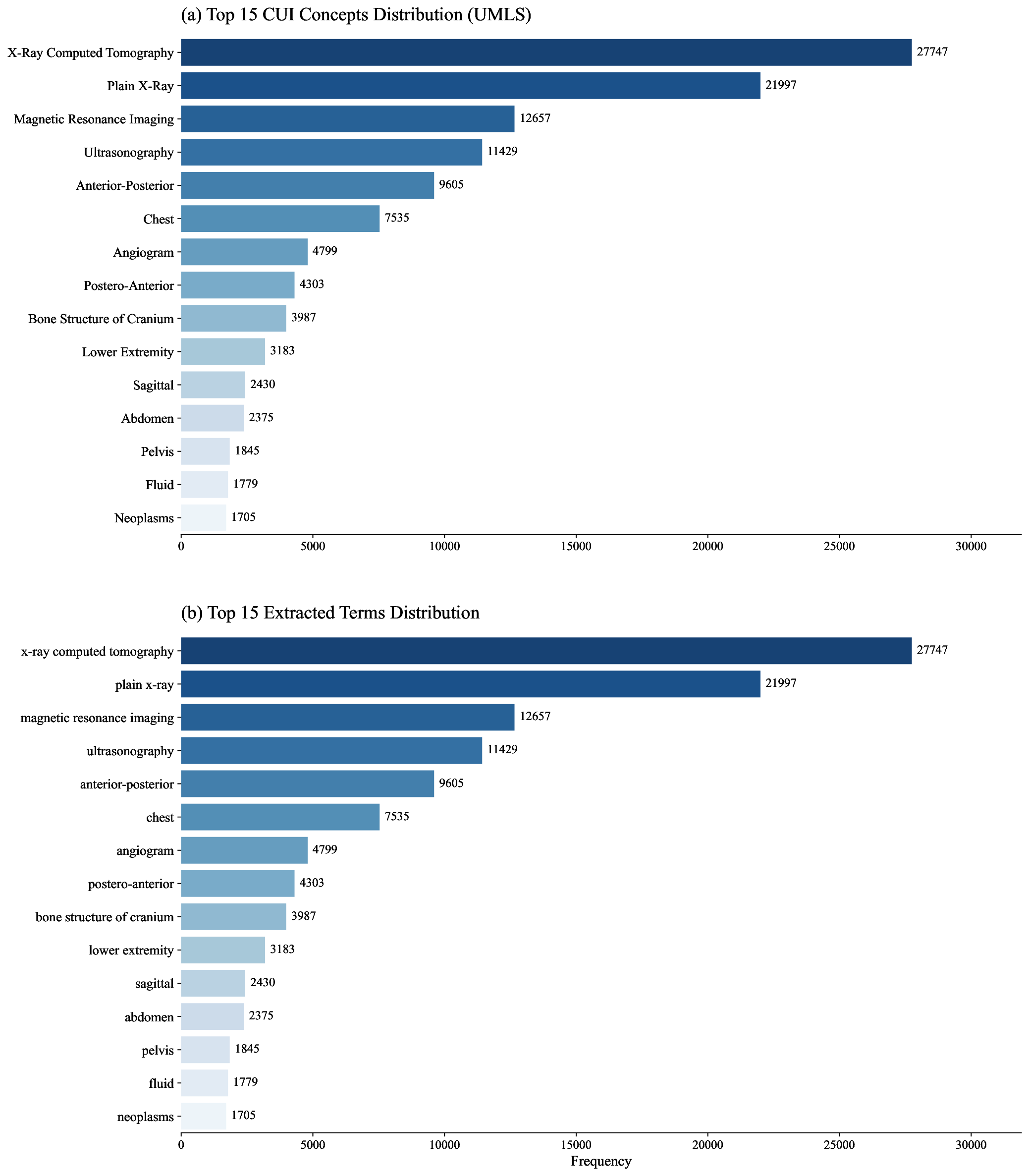}
    \caption{\textbf{Distribution of Top 15 Concepts and Terms.} 
    (a) The frequency of Unique Medical Language System~(UMLS) Concept Unique Identifiers~(CUIs) mapped from the dataset.
    (b) The frequency of raw extracted terms directly from the captions. 
    Both distributions illustrate the long-tail nature of medical findings in the dataset.}
    \label{fig:appendix_stats}
\end{figure*}

MedLayBench-V encompasses a diverse range of medical imaging modalities, mirroring real-world clinical distributions.
As summarized in Table~\ref{tab:roco_distribution}, Computed Tomography (CT) and X-Ray constitute the majority of the dataset, reflecting their prevalence in diagnostic radiology.
Table~\ref{tab:detailed_modality_stats} further breaks down the top co-occurring concepts for each modality, confirming that our extraction pipeline correctly identifies modality-specific anatomical structures (e.g., "left ventricle" in Ultrasound, "coronary artery" in Angiography).
Additionally, Figure~\ref{fig:appendix_stats} illustrates the long-tail distribution of both UMLS concepts and raw terms.
This indicates that while a few common concepts dominate the distribution~(head), the dataset also preserves a vast array of rare, specific medical conditions~(tail), which is crucial for comprehensive evaluation of medical VLMs.

\begin{table}[h!]
    \centering
    \captionsetup{labelfont=bf}
    \caption{\textbf{Distribution of Imaging Modalities.} The number of image-caption pairs for each modality as reported in the original ROCOv2 dataset \cite{ruckert2024rocov2}.}
    \label{tab:roco_distribution}
    \resizebox{\columnwidth}{!}{
    \setlength{\tabcolsep}{5pt}
    \renewcommand{\arraystretch}{1.0}
    \begin{tabular}{l l r}
        \toprule
        \textbf{Code} & \textbf{Modality Name} & \textbf{Count} \\
        \midrule
        DRCT & Computed Tomography (CT)       & 27,747 \\
        DRXR & X-Ray (Plain Radiography)      & 21,997 \\
        DRMR & Magnetic Resonance Imaging (MRI) & 12,657 \\
        DRUS & Ultrasonography                & 11,429 \\
        DRAN & Angiography                    & 4,799  \\
        DRCO & Combined Modality              & 728    \\
        DRPE & Positron Emission Tomography (PET) & 432 \\
        \midrule
        \textbf{Total} & & \textbf{79,789} \\
        \bottomrule
    \end{tabular}
    }
\end{table}

\begin{table*}[h]
    \centering
    \captionsetup{labelfont=bf}
    \caption{\textbf{Detailed Top 5 Concepts Distribution per Modality.} The frequency of the top 5 co-occurring concepts extracted from the text context for each major imaging modality.}
    \label{tab:detailed_modality_stats}
    
    \begin{minipage}[t]{0.48\textwidth}
        \centering
        \subcaption{\textbf{Computed Tomography (CT)}}
        \setlength{\tabcolsep}{5pt}
        \small
        \begin{tabular}{cl r}
            \toprule
            \textbf{Rank} & \textbf{Concept} & \textbf{Freq} \\
            \midrule
            1 & X-Ray Computed Tomography & 27,747 \\
            2 & CT scan & 3,474 \\
            3 & abdomen & 2,869 \\
            4 & arrow & 2,802 \\
            5 & image & 1,669 \\
            \bottomrule
        \end{tabular}
    \end{minipage}
    \hfill
    \begin{minipage}[t]{0.48\textwidth}
        \centering
        \subcaption{\textbf{Magnetic Resonance Imaging (MRI)}}
        \setlength{\tabcolsep}{5pt}
        \small
        \begin{tabular}{cl r}
            \toprule
            \textbf{Rank} & \textbf{Concept} & \textbf{Freq} \\
            \midrule
            1 & Magnetic Resonance Imaging & 12,659 \\
            2 & arrow & 1,246 \\
            3 & image & 957 \\
            4 & patient & 712 \\
            5 & brain & 661 \\
            \bottomrule
        \end{tabular}
    \end{minipage}
    
    \vspace{0.5cm} 
    
    \begin{minipage}[t]{0.48\textwidth}
        \centering
        \subcaption{\textbf{Ultrasonography}}
        \setlength{\tabcolsep}{5pt}
        \small
        \begin{tabular}{cl r}
            \toprule
            \textbf{Rank} & \textbf{Concept} & \textbf{Freq} \\
            \midrule
            1 & Ultrasonography & 11,422 \\
            2 & arrow & 892 \\
            3 & image & 633 \\
            4 & left ventricle & 610 \\
            5 & left atrium & 564 \\
            \bottomrule
        \end{tabular}
    \end{minipage}
    \hfill
    \begin{minipage}[t]{0.48\textwidth}
        \centering
        \subcaption{\textbf{Plain X-Ray}}
        \setlength{\tabcolsep}{5pt}
        \small
        \begin{tabular}{cl r}
            \toprule
            \textbf{Rank} & \textbf{Concept} & \textbf{Freq} \\
            \midrule
            1 & Plain x-ray & 21,936 \\
            2 & Anterior-Posterior (AP) & 9,606 \\
            3 & Chest & 7,196 \\
            4 & Postero-Anterior (PA) & 4,302 \\
            5 & Bone structure of cranium & 3,973 \\
            \bottomrule
        \end{tabular}
    \end{minipage}

    \vspace{0.5cm} 

    \begin{minipage}[t]{0.48\textwidth}
        \centering
        \subcaption{\textbf{Angiography}}
        \setlength{\tabcolsep}{5pt}
        \small
        \begin{tabular}{cl r}
            \toprule
            \textbf{Rank} & \textbf{Concept} & \textbf{Freq} \\
            \midrule
            1 & angiogram & 4,766 \\
            2 & arrow & 435 \\
            3 & Coronary angiography & 228 \\
            4 & right coronary artery & 219 \\
            5 & stenosis & 195 \\
            \bottomrule
        \end{tabular}
    \end{minipage}
    \hfill
    \begin{minipage}[t]{0.48\textwidth}
        \centering
        \subcaption{\textbf{Positron-Emission Tomography (PET)}}
        \setlength{\tabcolsep}{5pt}
        \small
        \begin{tabular}{cl r}
            \toprule
            \textbf{Rank} & \textbf{Concept} & \textbf{Freq} \\
            \midrule
            1 & Positron-Emission Tomography & 432 \\
            2 & uptake & 98 \\
            3 & increased & 64 \\
            4 & image & 40 \\
            5 & patient & 34 \\
            \bottomrule
        \end{tabular}
    \end{minipage}
\end{table*}

\section{Semantic Preservation Analysis}
\label{sec:appendix_semantic}

\begin{figure*}[h]
    \centering
    \includegraphics[width=\textwidth]{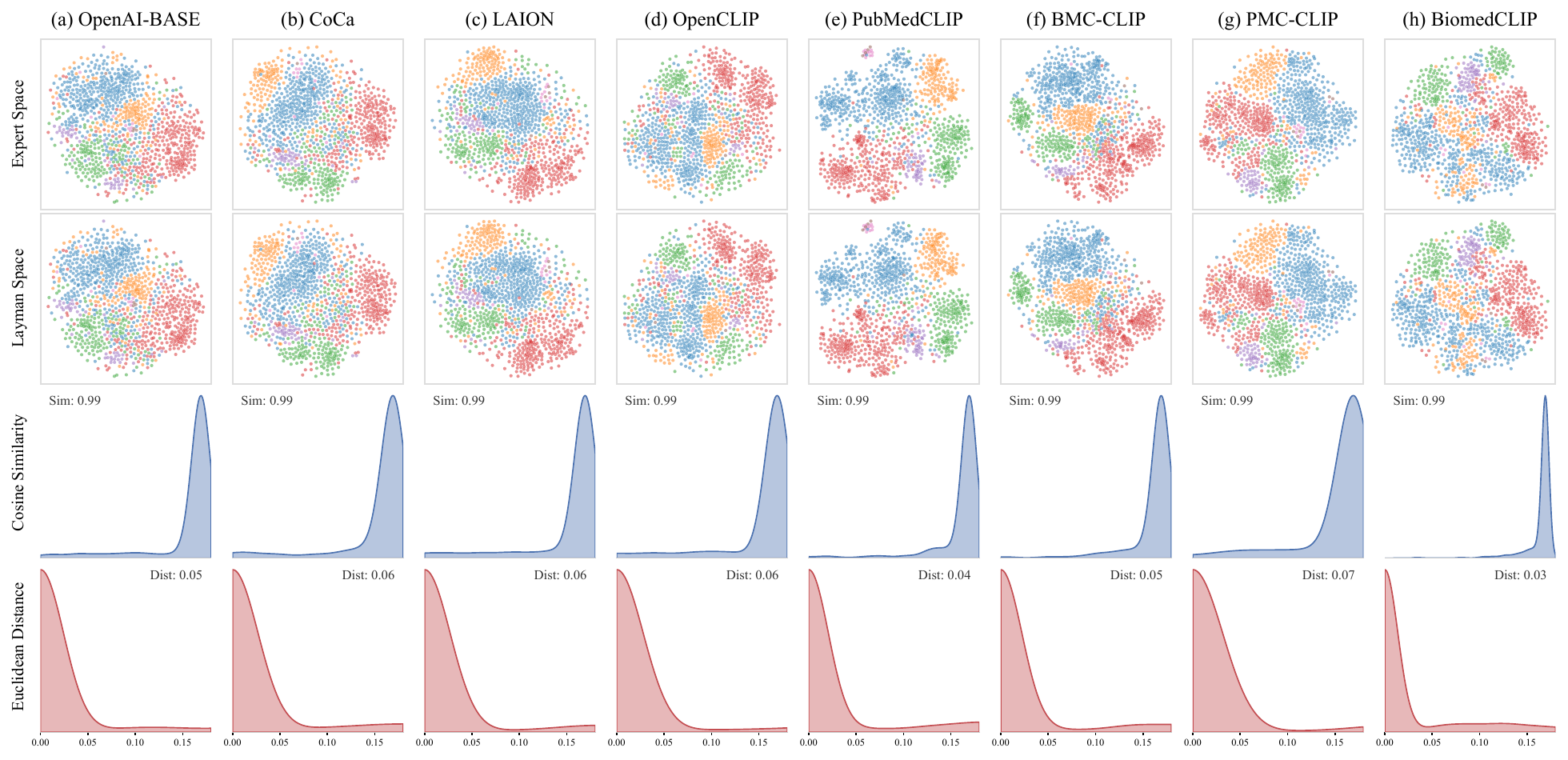}
    \caption{
        \textbf{Embedding space visualization across different CLIP models.}
        Each column represents a different model.
        Rows 1--2: t-SNE projections of Expert and Layman embeddings, colored by modality.
        Row 3: Cosine similarity distribution. Row 4: Euclidean distance distribution.
        High similarity (Sim $\approx$ 0.99) and low distance (Dist $\approx$ 0.05--0.07) confirm semantic preservation across all models.
    }
    \label{fig:embedding_analysis}
\end{figure*}

To empirically validate that our simplification process preserves the underlying medical semantics, we analyzed the embedding space of various Vision-Language Models.
Figure~\ref{fig:embedding_analysis} visualizes the t-SNE projections of image-text embeddings for both Expert (original) and Layman (refined) captions.
Across different architectures (OpenAI-CLIP, BiomedCLIP, PMC-CLIP), we observe that the distributions of Expert and Layman embeddings are nearly isomorphic.
Furthermore, the high cosine similarity ($\approx 0.99$) and low Euclidean distance distributions confirm that the transition to lay language does not shift the semantic vector significantly.
This serves as strong evidence that MedLayBench-V successfully lowers the linguistic barrier without compromising the diagnostic information required for downstream evaluation.

\section{Bootstrap Significance Test}
\label{sec:appendix_bootstrap}

To verify that the performance differences between Expert and Layman captions are not attributable to sampling variance, we conducted bootstrap significance testing ($n$=1{,}000, two-sided) on the Overall Recall@$K$ delta, as summarized in Table~\ref{tab:bootstrap}.

General-domain models show largely non-significant differences ($p > .05$), consistent with their low baseline where minor fluctuations are indistinguishable from noise.
Medical-domain models exhibit statistically significant drops ($p < .05$) across all metrics, with the largest delta observed for BiomedCLIP at R@10 ($-$0.75\%).
Nevertheless, all $|\Delta|$ remain below 1.03\%, confirming the degradation is statistically detectable but practically negligible for retrieval.

\begin{table}[ht]
\centering
\caption{\textbf{Bootstrap Significance Test.} 
$\Delta$ denotes Layman $-$ Expert (\%). * indicates $p < .05$.}
\label{tab:bootstrap}
\resizebox{\columnwidth}{!}{
\setlength{\tabcolsep}{4pt}
\renewcommand{\arraystretch}{1.2}
\begin{tabular}{l|cc|cc|cc}
\toprule
\multirow{2}{*}{\textbf{Model}} & \multicolumn{2}{c|}{\textbf{R@1}} & \multicolumn{2}{c|}{\textbf{R@5}} & \multicolumn{2}{c}{\textbf{R@10}} \\
\cmidrule(lr){2-3} \cmidrule(lr){4-5} \cmidrule(lr){6-7}
 & $\Delta$ & $p$ & $\Delta$ & $p$ & $\Delta$ & $p$ \\
\midrule
\multicolumn{7}{l}{\textit{\textbf{General-Domain Models}}} \\
\quad OpenAI & $-$0.07 & .020* & $-$0.07 & .222 & $-$0.13 & .148 \\
\quad CoCa & $-$0.03 & .590 & $-$0.08 & .188 & $-$0.12 & .056 \\
\quad LAION & $-$0.04 & .404 & $-$0.05 & .248 & $-$0.22 & .032* \\
\quad OpenCLIP & +0.04 & .980 & $-$0.13 & .202 & $-$0.16 & .076 \\
\midrule
\multicolumn{7}{l}{\textit{\textbf{Medical-Domain Models}}} \\
\quad PubMedCLIP & $-$0.20 & .006* & $-$0.28 & .004* & $-$0.32 & .008* \\
\quad BMC-CLIP & $-$0.10 & .036* & $-$0.45 & .002* & $-$0.57 & .001* \\
\quad PMC-CLIP & $-$0.66 & .001* & $-$0.67 & .001* & $-$0.64 & .001* \\
\quad BiomedCLIP & $-$0.41 & .001* & $-$0.65 & .001* & $-$0.75 & .001* \\
\bottomrule
\end{tabular}
}
\end{table}

\begin{table*}[h]
\centering
\caption{\textbf{Per-Model SCGR Ablation Results.}
Averaged R@1 (\%) across I2T and T2I for each ablation condition. Condition definitions follow Table~\ref{tab:ablation_scgr}.}
\label{tab:ablation_full}
\setlength{\tabcolsep}{2pt}
\renewcommand{\arraystretch}{1.2}
\begin{tabular}{l|ccc|cccc|cccc|c}
\toprule
\multirow{2}{*}{\textbf{Condition}} & \multicolumn{3}{c|}{\textbf{Pipeline Steps}} & \multicolumn{4}{c|}{\textbf{General-Domain}} & \multicolumn{4}{c|}{\textbf{Medical-Domain}} & \multirow{2}{*}{\textbf{Avg.}} \\
\cmidrule(lr){2-4} \cmidrule(lr){5-8} \cmidrule(lr){9-12}
 & CUI & MedLP & LLM & OpenAI & CoCa & LAION & CLIP & PubMed & BMC & PMC & BioMed & \\
\midrule
LLM & $\times$ & $\times$ & \cmark & 0.50 & 0.77 & 0.74 & 1.03 & 1.18 & 3.32 & 4.19 & 3.96 & 1.96 \\
CUI & \cmark & $\times$ & \cmark & 0.41 & 0.78 & 0.93 & 1.17 & 1.24 & 3.58 & 4.51 & 4.00 & 2.08 \\
SCGR (Ours) & \cmark & \cmark & \cmark & 0.99 & 2.08 & 2.52 & 3.38 & 3.99 & 20.53 & 27.27 & \textbf{29.30} & \textbf{11.26} \\
\midrule
Expert & -- & -- & -- & 1.06 & 2.11 & 2.56 & 3.34 & 4.19 & 20.63 & 27.93 & 29.71 & 11.44 \\
\bottomrule
\end{tabular}
\end{table*}

\section{Ablation Study}
\label{sec:ablation_llm_only}
\label{sec:appendix_ablation_full}

Table~\ref{tab:ablation_full} extends the averaged ablation results in Table~\ref{tab:ablation_scgr} with a per-model breakdown.
The LLM Only condition shows uniformly poor performance across all models, with medical-domain models suffering disproportionately larger gaps relative to Expert.
Adding CUI extraction alone provides marginal gains, confirming that ontological grounding is insufficient without lexical substitution via MedlinePlus.
Full SCGR recovers near-Expert performance consistently, with BiomedCLIP showing the largest absolute improvement from 3.96 to 29.30.

\begin{figure*}[h!]
    \centering
    \renewcommand{\thesubfigure}{\roman{subfigure}}
    \begin{subfigure}[t]{\textwidth}
        \centering
        \includegraphics[width=\textwidth]{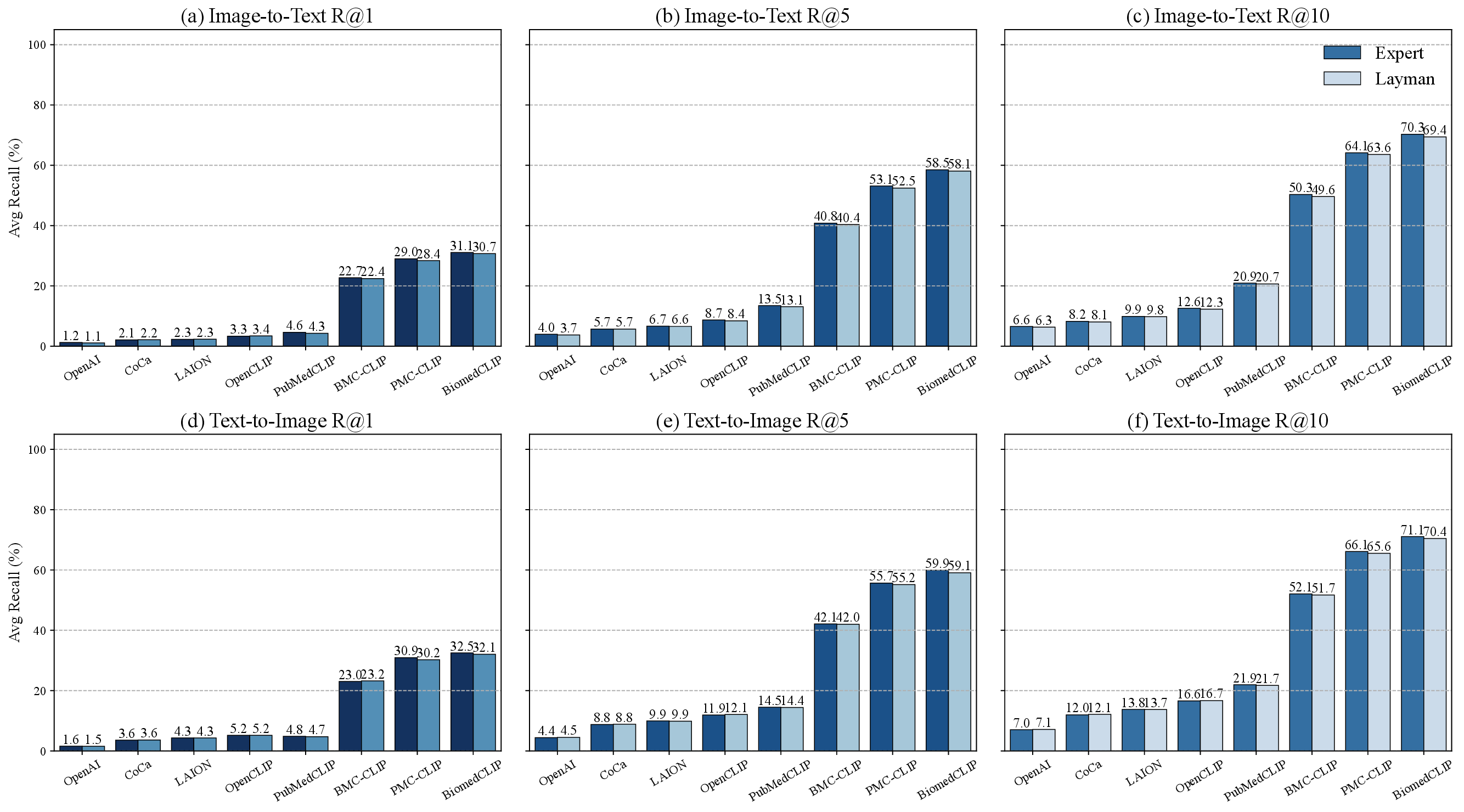}
        \caption{\textbf{Zero-Shot Retrieval Performance.} 
        Recall@$K$ results for Image-to-Text (a--c) and Text-to-Image (d--f) tasks. 
        Dark and light bars denote Expert and Layman queries, respectively.}
        \label{fig:retrieval_graph}
    \end{subfigure}
    \vspace{0.3cm}
    \begin{subfigure}[t]{\textwidth}
        \centering
        \includegraphics[width=\textwidth]{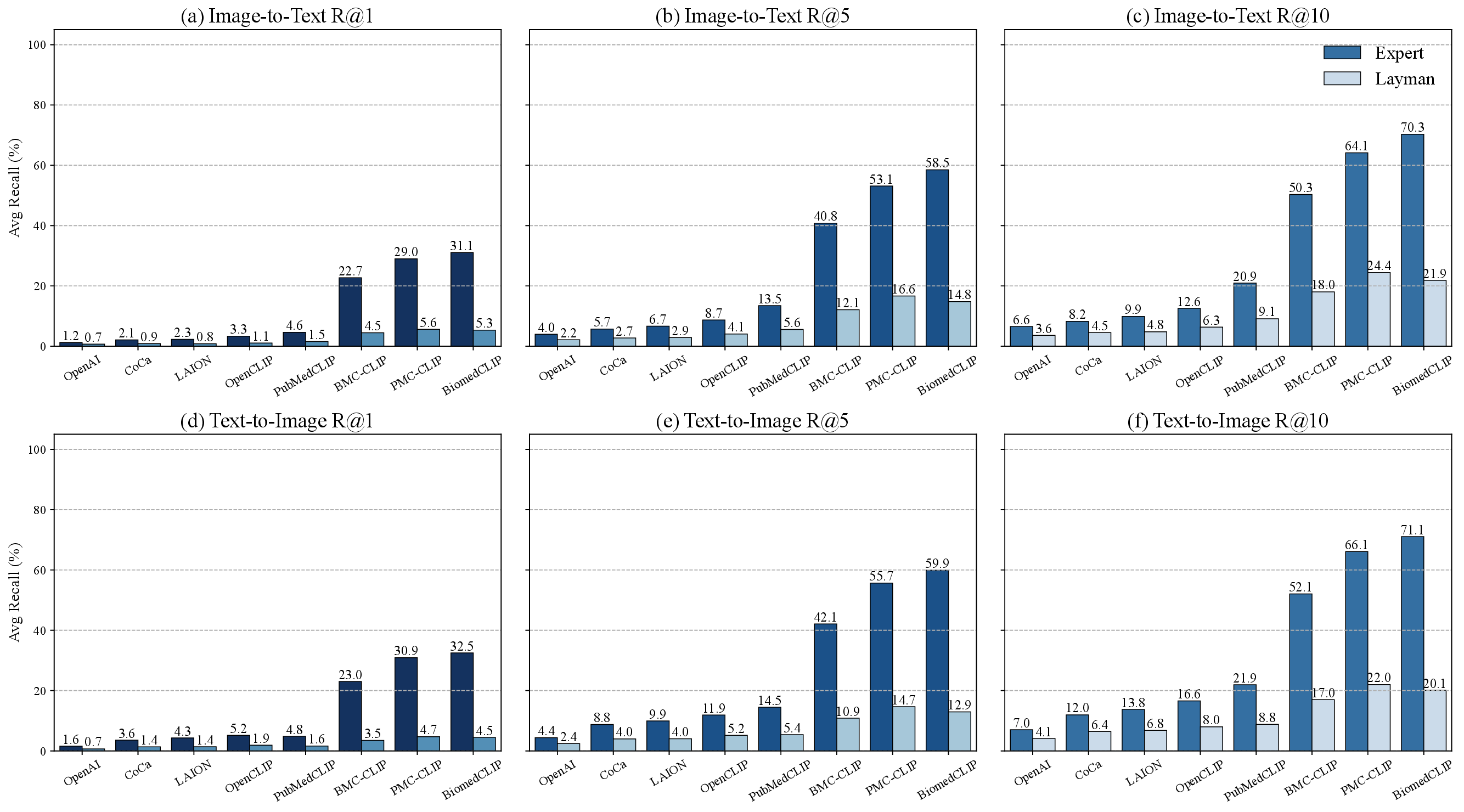}
        \caption{\textbf{Impact of Naive LLM-only Simplification.} 
        Recall@$K$ results using layman captions generated without structured grounding.
        BiomedCLIP I2T R@1 collapses from 31.1\% to 5.3\%.}
        \label{fig:ablation_llm_only}
    \end{subfigure}
    \caption{\textbf{Retrieval Performance and Ablation Visualization.}
    (i) SCGR preserves semantic fidelity with negligible gaps between registers.
    (ii) Naive LLM simplification causes severe semantic drift across all models.}
    \label{fig:retrieval_ablation}
\end{figure*}

Figure~\ref{fig:retrieval_graph} visualizes the retrieval performance across all models, confirming negligible gaps between Expert and SCGR-generated Layman captions.
In contrast, Figure~\ref{fig:ablation_llm_only} illustrates that removing structured grounding causes severe degradation, with BiomedCLIP I2T R@1 collapsing from 31.1\% to 5.3\%.
We identify two dominant failure modes of the Naive LLM.
First, it tends to over-simplify specific pathologies into vague terms (e.g., ``pneumothorax'' $\rightarrow$ ``lung problem''), losing discriminative features.
Second, it hallucinates plausible but incorrect details to fill narrative gaps.
These findings confirm that explicit knowledge grounding, as provided by SCGR, is essential for high-quality medical lay language generation.



\renewcommand{\thesubfigure}{\alph{subfigure}} 

\section{Zero-Shot Captioning Analysis}
\label{sec:appendix_captioning}

To complement the retrieval-based evaluation, we conducted a zero-shot image captioning experiment to directly assess whether current VLMs can adapt their output register.
Two medical-domain models, LLaVA-Med~\cite{li2023llava} and MedGemma~1.5~\cite{sellergren2025medgemma}, and two general-domain models, LLaVA-v1.5~\cite{liu2024improved} and Qwen2-VL~\cite{wang2024qwen2}, each received dual prompts per image on 1{,}000 test pairs: (A) ``Describe this medical image in one sentence using clinical terminology'' and (B) ``Describe this medical image in one sentence using simple language that a patient with no medical background can understand.''
We report BERTScore~\cite{zhang2019bertscore} against Expert and Layman references respectively, along with FKGL to measure readability shift.

\begin{table}[h]
\centering
\caption{\textbf{Zero-Shot Captioning Results.}
BERTScore (DeBERTa-xlarge-MNLI) against register-matched references. $\Delta$ = Expert $-$ Layman; positive indicates expert-register bias.}
\label{tab:captioning}
\resizebox{\columnwidth}{!}{
\setlength{\tabcolsep}{4pt}
\renewcommand{\arraystretch}{1.2}
\begin{tabular}{l|ccc|cc}
\toprule
\textbf{Model} & \textbf{BS\textsubscript{Exp}} & \textbf{BS\textsubscript{Lay}} & \textbf{$\Delta$} & \textbf{FKGL\textsubscript{Exp}} & \textbf{FKGL\textsubscript{Lay}} \\
\midrule
\multicolumn{6}{l}{\textit{\textbf{Medical-Domain}}} \\
LLaVA-Med & 55.04 & 32.12 & \textbf{+22.93} & 7.2 & 4.1 \\
MedGemma 1.5 & 64.31 & 65.11 & $-$0.80 & 13.6 & 6.5 \\
\midrule
\multicolumn{6}{l}{\textit{\textbf{General-Domain}}} \\
LLaVA-v1.5 & 61.13 & 63.05 & $-$1.92 & 4.8 & 5.0 \\
Qwen2-VL & 63.05 & 65.28 & $-$2.23 & 15.0 & 9.0 \\
\bottomrule
\end{tabular}
}
\end{table}

As shown in Table~\ref{tab:captioning}, LLaVA-Med~\cite{li2023llava} shows a severe expert bias ($\Delta$=+22.93) despite producing syntactically simpler outputs (FKGL 7.2$\rightarrow$4.1), indicating the bottleneck lies in vocabulary register rather than syntactic complexity.
The remaining models exhibit near-zero gaps ($\Delta$=$-$0.80 to $-$2.23) with notable readability shifts, suggesting that lay-register adaptability varies across VLM families.
This heterogeneity motivates the need for a standardized benchmark like MedLayBench-V to systematically evaluate and improve expert-lay alignment.

\section{Extended Qualitative Analysis}
\label{sec:appendix_qualitative}

\begin{figure*}[hbt!]
    \centering
    \includegraphics[width=0.9\textwidth]{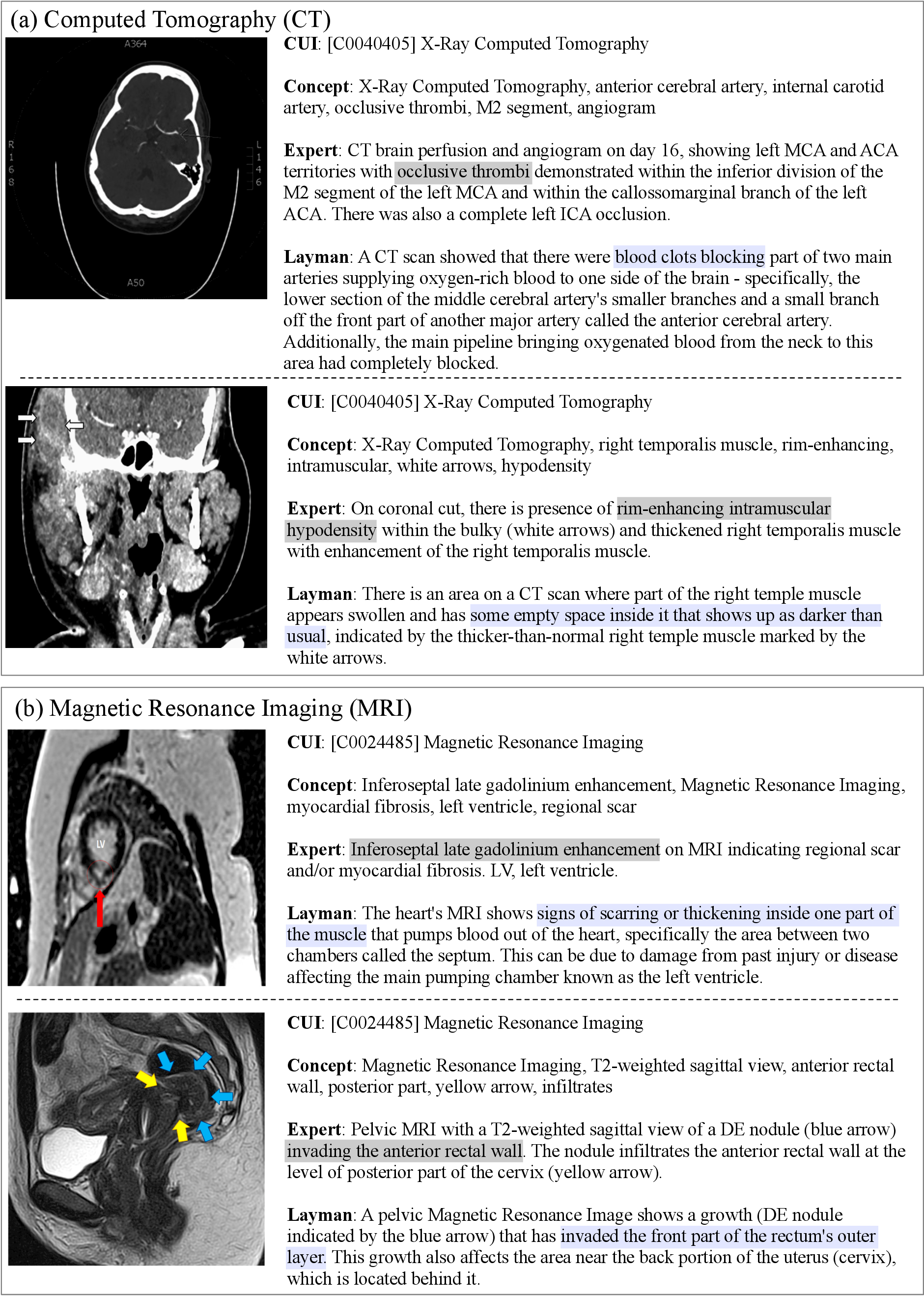}
    \caption{
    \textbf{Qualitative Analysis on Cross-Sectional Modalities.}
    Comparison of expert and layman descriptions for (a) CT and (b) MRI.
    }
    \label{fig:quali_part1}
\end{figure*}

\begin{figure*}[hbt!]
    \centering
    \includegraphics[width=0.9\textwidth]{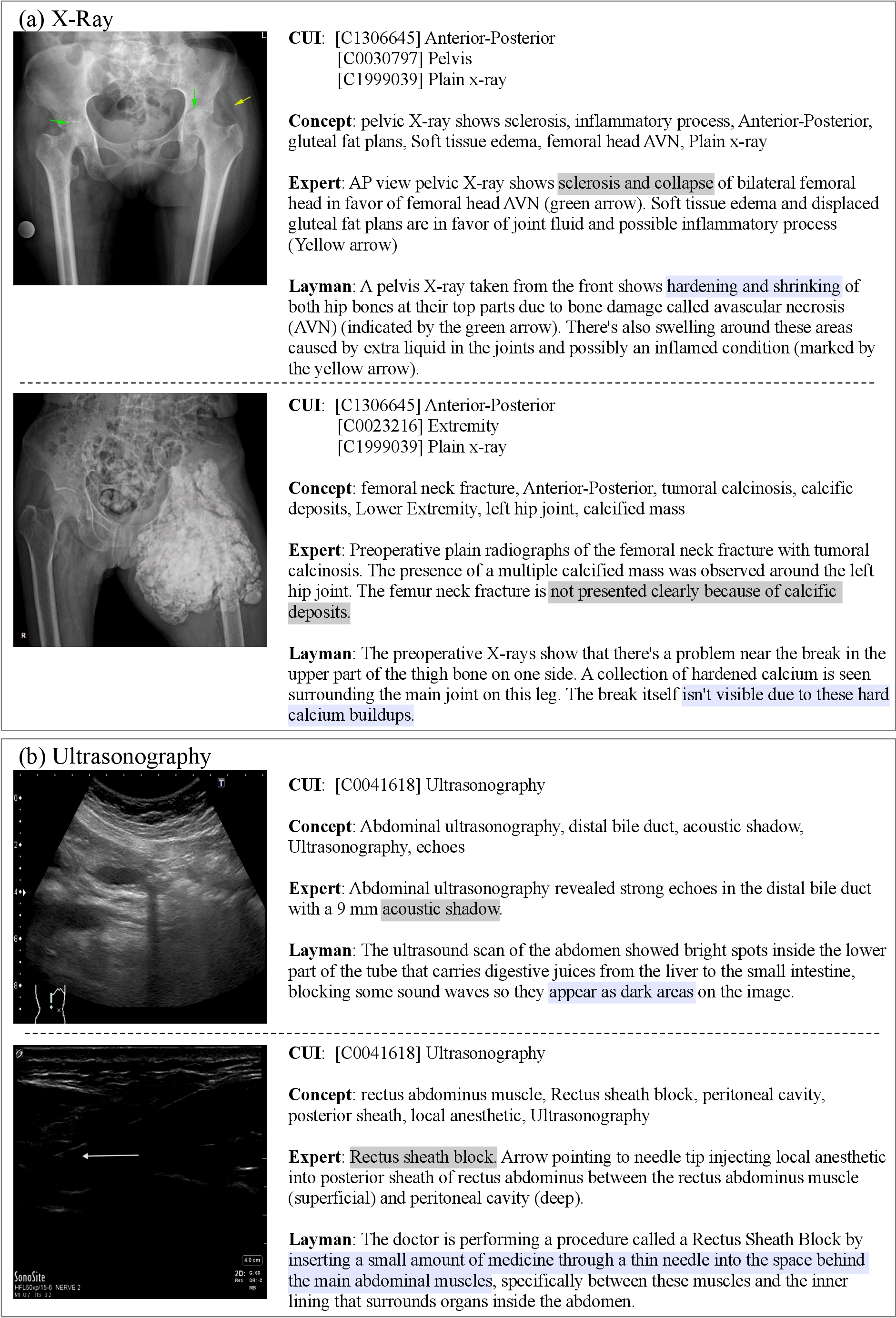}
    \caption{
    \textbf{Qualitative Analysis on Cross-Sectional Modalities.}
    Comparison of expert and layman descriptions for (a) X-Ray and (b) Ultrasonography.
    }
    \label{fig:quali_part2}
\end{figure*}

To further demonstrate the robustness and versatility of the SCGR pipeline, we provide an extended set of qualitative examples across diverse imaging modalities.
Figure~\ref{fig:quali_part1} and Figure~\ref{fig:quali_part2} illustrate how our pipeline handles specific linguistic challenges, ranging from simplifying complex vascular anatomy in CT/MRI to interpreting acoustic artifacts in ultrasound.
Each example highlights the transformation from the original expert report (Expert) to the generated patient-friendly caption (Layman).
Key medical terms are highlighted in grey while their simplified explanations are highlighted in blue to visualize the semantic alignment.

\end{document}